\documentclass[journal]{IEEEtran}
\ifCLASSINFOpdf
  \usepackage[pdftex]{graphicx}
  \graphicspath{{../pdf/}{../jpeg/}}
  \DeclareGraphicsExtensions{.pdf,.jpeg,.png,.jpg}
\else

\fi

\usepackage[cmex10]{amsmath}
\usepackage{color}

\usepackage{amssymb}

\usepackage{multirow}

\begin{document}
%
\title{Leveraging Virtual and Real Person for Unsupervised Person Re-identification}
%
%
%

\author{Fengxiang Yang,
        Zhun Zhong,
        Zhiming Luo,
        Sheng Lian,
        and Shaozi Li
\thanks{F. Yang, Z. Zhong, S. Lian and S. Li are with the Department of Cognitive Science, Xiamen University, Xiamen, 361005, China}
\thanks{Z. Luo is with Postdoc Center of Information and Communication Engineering, Xiamen University, Xiamen, 361005, China.}
}

\markboth{Journal of \LaTeX\ Class Files,~Vol.~14, No.~8, November~2018}%
{Shell \MakeLowercase{\textit{et al.}}: Bare Demo of IEEEtran.cls for IEEE Journals}

\maketitle
\begin{abstract}
Person re-identification (re-ID) is a challenging problem especially when no labels are available for training. Although recent deep re-ID methods have achieved great improvement, it is still difficult to optimize deep re-ID model without annotations in training data. To address this problem, this study introduces a novel approach for unsupervised person re-ID by leveraging virtual and real data. Our approach includes two components: \textit{virtual person generation} and \textit{training of deep re-ID model}. For virtual person generation, we learn a person generation model and a camera style transfer model using unlabeled real data to generate virtual persons with different poses and camera styles. 
The virtual data is formed as labeled training data, enabling subsequently training deep re-ID model in supervision. For training of deep re-ID model, we divide it into three steps: 1) pre-training a coarse re-ID model by using virtual data; 2) collaborative filtering based positive pair mining from the real data; and 3) fine-tuning of the coarse re-ID model by leveraging the mined positive pairs and virtual data. The final re-ID model is achieved by iterating between step 2 and step 3 until convergence. Experimental results on two large-scale datasets, Market-1501 and DukeMTMC-reID, demonstrate the effectiveness of our approach and shows that the state of the art is achieved in unsupervised person re-ID.
\end{abstract}

\begin{IEEEkeywords}
Person re-ID, Generative Adversarial Network, Collaborative Filtering.
\end{IEEEkeywords}

\IEEEpeerreviewmaketitle

\section{Introduction}
\label{sec:intro}
Person re-identification (re-ID) aims to find the same person from a gallery collected by different cameras. It is a challenging task due to the large image variations caused by different human poses and camera settings. During the past few years, person re-ID has achieved great improvement, benefiting from the remarkable success of deep Convolutional Neural Nets (CNNs) \cite{krizhevsky2012alex,he2016deep}. Nevertheless, training deep re-ID model requires a substantial annotated data, which is quite expensive especially when across a mass of cameras. Under such circumstances, there is an urgent demand for learning the discriminative deep re-ID model with large-scale unlabeled data. In this paper, we address the challenging unsupervised person re-ID problem, where large-scale training data is provided while no label information is available.

Unsupervised person re-ID has been studied in many previous works \cite{liao2015person,yang2017unsupervised,zheng2015scalable}. These works mainly focus on designing discriminative hand-crafted features and dealing with a small dataset, but degenerate when applying on large-scale datasets. Deep CNNs have reached state-of-the-art performance on large-scale person re-ID datasets. Most of existing deep CNNs based re-ID models were trained by using either ID-discriminative embedding (IDE) \cite{zheng2016person} or triplet (or pairwise) loss \cite{HermansBeyer2017Arxiv}. However, it is impossible to train these models without annotations on the training set, because both IDE and triplet loss require label information or the relationship (positive and negative) with other training data for the given image. There are limited works make efforts on deep learning based unsupervised re-ID. Fan \textit{et al.} \cite{fan2017unsupervised} propose a framework called UPL which progressively utilizes $k$-means clustering to find reliable positive pairs and fine-tunes the deep CNN model. The main drawbacks of UPL are that the initial re-ID model should be pre-trained on a labeled re-ID dataset and the rough number of unique identities in the target dataset should be given for clustering.

\begin{figure*}[!t]
  \centering
  \includegraphics[width=\linewidth]{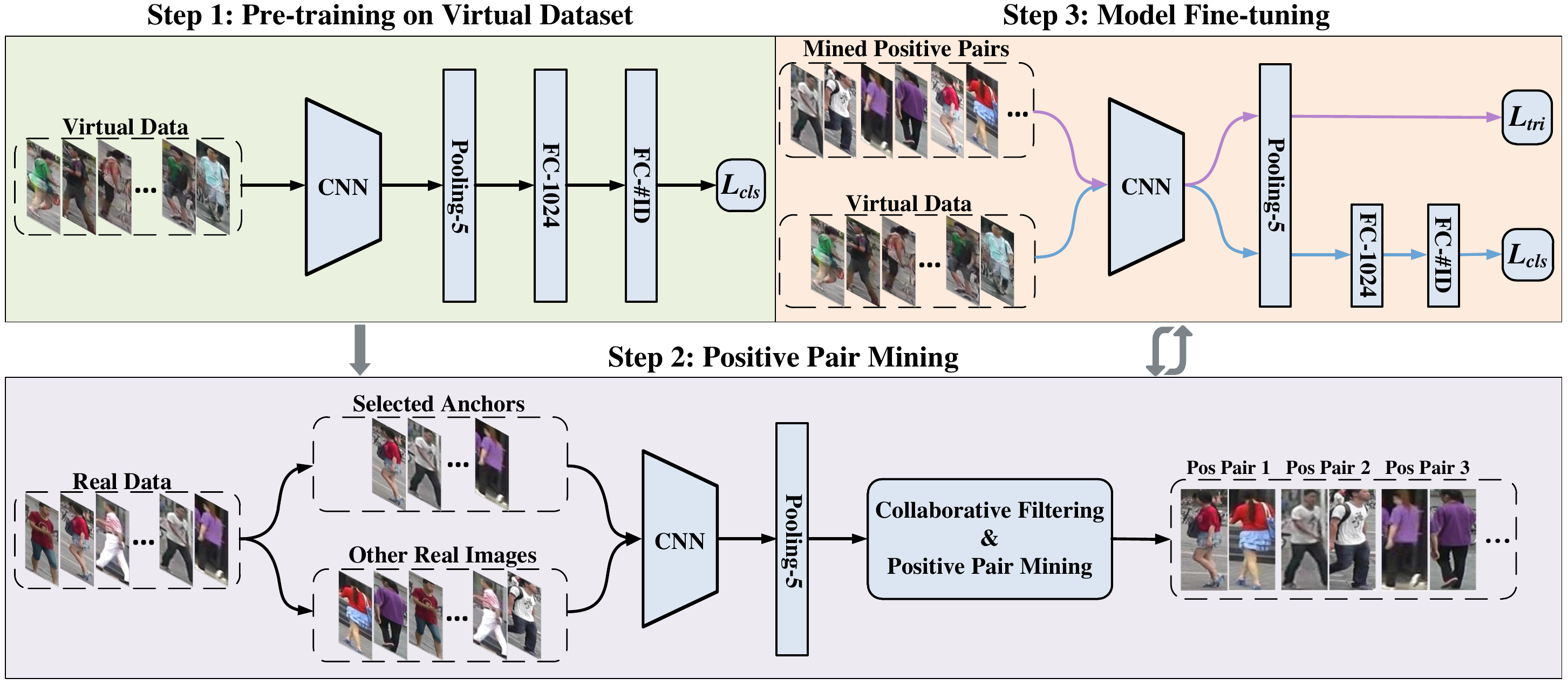}
   \caption{The overall framework of the training of deep re-ID model which can be divided into three steps. In step 1, we use virtual data generated by DPG-GAN and Star-GAN to train a coarse deep re-ID model. Then, a collaborative filtering based positive pair mining approach is utilized to find reliable positive pairs from the real data in step 2. In step 3, we refine the coarse re-ID model by leveraging the virtual data and mined positive pairs with a multi-task loss function. Finally, we alternate between step 2 and step 3 until the re-ID model converged.}
  \label{fig:process}
\end{figure*}

In this study, we propose a deep CNN based approach for unsupervised person re-ID. Our approach consists of two components: 1) virtual person generation and 2) training of deep re-ID model. For virtual person generation, we first employ DPG-GAN \cite{ma2018disentangled} and Star-GAN \cite{choi2017stargan} to learn a person generation model and a camera style transfer model by using unlabeled real training data. As such, we can generate virtual persons with different poses and assign them with corresponding pseudo labels. Then the same generated identity will be style transfered to different cameras. These virtual persons are formed as virtual training data and subsequently be utilized to train a coarse deep re-ID model in a supervised way.

For training of deep re-ID model, it is divided into three steps as shown in Figure~\ref{fig:process}: 1) pre-training on virtual data, 2) positive pair mining, and 3) model fine-tuning. 

For step 1, a coarse deep re-ID model is trained by using the generated virtual data. This model can provide discriminative representation for measuring the similarity of persons. 

For step 2, we first use the previous pre-trained coarse re-ID model to extract features for each real image and compute its $k$-reciprocal nearest neighbors ($k$-RNNs) \cite{qin2011hello}. Although each image and one of its $k$-RNNs can be treated as a positive pair, there are a large amount of false positive pairs which have negative effects for fine-tuning the re-ID model. In order to alleviate this issue, a novel collaborative filtering based positive pair mining approach is proposed to find most reliable positive pairs on unlabeled real data.

In step 3, the mined positive pairs along with the virtual labeled training data are simultaneously leveraged for model refinement by using a multi-task loss function in which IDE for virtual data and triplet loss for mined positive pairs. At last, the final deep re-ID model is achieved by iterating between step 2 and step 3 until convergence. 

To summarize, the main contributions of this study are as follows:
\begin{itemize}
    \item We propose a novel framework for unsupervised person re-ID by leveraging the generated pseudo labeled virtual data and the unlabeled real data for deep re-ID network training.
	\item A collaborative filtering based positive pair mining approach is used to select reliable training pairs from unlabeled real data for model refinement by leveraging person-to-person similarity relations.
    \item The proposed method achieves the state-of-the-art performance in unsupervised person re-ID on two large-scale datasets, Market-1501 and DukeMTMC-reID.
\end{itemize}

\section{Related Work}
\label{sec:review}
\textbf{Unsupervised Person Re-identification.} Unsupervised person re-ID attempts to learn discriminate features for pedestrian with unlabeled data. Hand-craft features can be directly employed for unsupervised person re-ID. Farenzena \textit{et al.} \cite{farenzena2010person} propose to use weighted color histogram, maximally stable color regions and recurrent high structured patches to separate foreground of pedestrians from background and compute appearance based feature for re-ID. Gray and Tao \cite{GrayTao} split input image into horizontal stripes and use 8 color channels and 21 texture filters on the luminance channel to extract feature. Recently, Zhao \textit{et al.} \cite{zhao2013unsupervised,zhao2013person,zhao2014learning} propose to split images of pedestrians into 10$\times$10 patches and combine LAB color histogram and SIFT feature as the final descriptor. Liao \textit{et al.} \cite{liao2015person} introduce local maximal occurrence descriptor (LOMO) by combining color feature and SILTP histogram. Zheng \textit{et al.} \cite{zheng2015scalable} propose to extract global visual feature by aggregating local color name descriptor. Bag of words model is then utilized for re-ID. Yang \textit{et al.} \cite{yang2017unsupervised} propose a weighted linear coding method for multi-level descriptor learning. These hand-craft based methods can be readily applied to unsupervised person re-ID but often fail to perform well on large-scale datasets.

Yu \textit{et al.} \cite{camel} present an unsupervised metric learning approach for re-ID called CAMEL. It employs asymmetric metric learning to find the shared space where the data representations are less affected by view-specific bias. Liu \textit{et al.} \cite{liu2017stepwise} propose a step-wise metric promotion model for unsupervised video person re-ID by iteratively estimating the annotations of training tracklets and optimizing re-ID model.

Recently, many works try to transfer a pre-trained re-ID model to the unlabeled dataset (also called domain adaptation). Peng \textit{et al.} \cite{peng2016unsupervised} exploit multi-task dictionary learning method to learn shared feature space between labeled dataset and unlabeled dataset. To take advantage of the strong discriminate ability of deep learning, Fan \textit{et al.} present a deep learning framework called UPL \cite{fan2017unsupervised}. They use a labeled dataset to initialize feature embedding and then fine-tune the network with positive sample pairs obtained through $k$-means clustering on unlabeled dataset. TJ-AIDL \cite{wang2018transferable} adopts a multi-branch network to establish an identity-discriminative and attribute-sensitive feature representation space most optimal for target domain without any label information. Deng \textit{et al.} \cite{deng2018image} introduce SP-GAN by jointly preserving self-similarity and domain-dissimilarity in the process of image-to-image translation. The source set is transferred to the style of target set and is then used to learn re-ID model for target set. Similarity, Wei \textit{et al.} \cite{wei2017person} present PT-GAN to reduce the domain gap by translating the given image to the style of target dataset and train deep re-ID model in supervised way. All the methods mentioned above require a labeled re-ID dataset to pre-train a re-ID model and then transfer it to the unlabeled target set. In this paper, we conduct unsupervised person re-ID under a more strict condition where there are only provided with unlabeled target set.

\textbf{Person Image Generation.} Generating realistic person images is a challenging task because of the complexity of foreground, person pose and background. The image generation models, \textit{e.g.}, VAE \cite{kingma2014auto} and GANs \cite{goodfellow2014generative}, have been demonstrated the effectiveness in person generation. Zhao \textit{et al.} \cite{zhao2017multi} combine variational inference into GAN to generate multi-view images of persons in a coarse-to-fine manner. Ma \textit{et al.} \cite{ma2017pose} develop a framework to generate new person images in arbitrary poses given as input person images and a target pose. Despite the promising results, these two approaches require aligned person image pairs in the training stage. To solve this problem, Esser \textit{et al.} \cite{esser2018VAE} propose VAE-U-Net to train person generation model by disentangling the shape and appearance of the input image. The novel image is generated with U-Net for target shape, conditioned on the VAE output for appearance. Ma \textit{et al.} \cite{ma2018disentangled} introduce DPG-GAN to generate virtual person images by simultaneously disentangle and encode the foreground, background and pose information into embedding features. The embedding features are then combined to reconstruct the input person image.

\textbf{Style Transfer.} Style transfer is a sub-domain of image-to-image translation. Recent works conducted on GANs \cite{goodfellow2014generative} have achieved impressive results on image-to-image translation. 
Pix2pix towards this goal by optimizing both adversarial and L1 loss of cGAN \cite{mirza2014cGAN}. However, paired samples are required in training process, this limits the application of pix2pix in practice. To alleviate this problem, Cycle-GAN \cite{zhu2017CycleGAN} introduces cycle-consistent loss to preserve key attributes for both source domain and target domain. These two models can only transfer images from one domain to another and may not be flexible enough when dealing with multi-domain translation. To overcome this problem, Star-GAN \cite{choi2017stargan} is proposed to combine classification loss and adversarial loss into training process to translate image into different styles with only one model.

\section{The Proposed Method}
\label{sec:model}

In this section, we first describe the pipeline of virtual person generation in Section~\ref{sec:person generation}. Then the implementing of training coarse Deep Re-ID model in Section~\ref{sec:fake-reid}. We present the details of collaborative filtering based positive pair mining in Section~\ref{sec:final-reid} and the final model fine-tuning in Section~\ref{sec:model-finetune}.

\subsection{Virtual Person Generation}
\label{sec:person generation}

In unsupervised person re-ID, identity annotations are not available in training set. This makes it difficult to train deep re-ID model in traditional way, such as IDE \cite{zheng2016person} and triplet loss \cite{HermansBeyer2017Arxiv}. To solve this problem, this paper considers to learn the potential distribution of the unlabeled person data, and to generate labeled virtual person images for deep re-ID model training. To achieve this goal, this work employs DPG-GAN \cite{ma2018disentangled} to generate person samples with different poses and transfer them to styles of different cameras by Star-GAN \cite{choi2017stargan}.

\begin{figure}
  \centering
  \includegraphics[width=\linewidth]{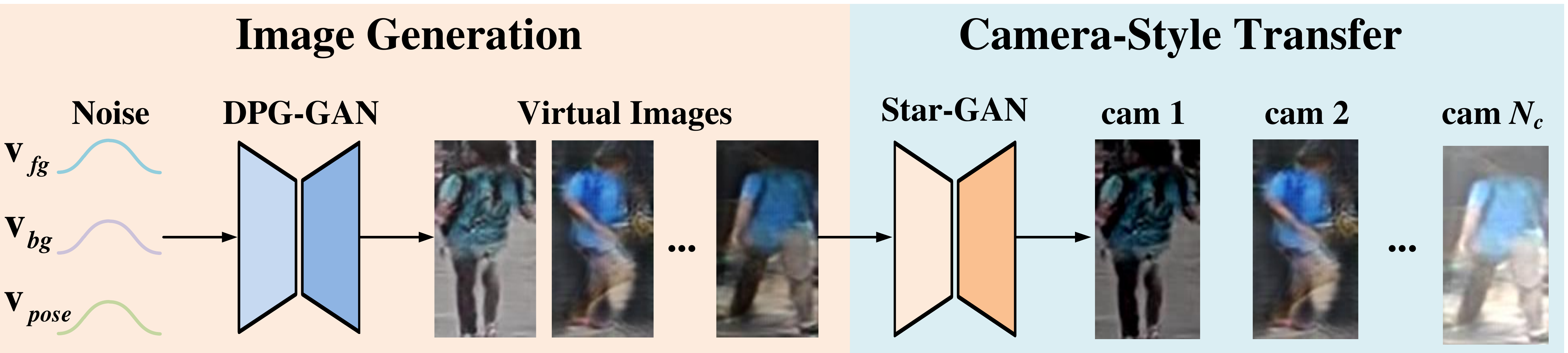}
  \caption{The pipeline of virtual image generation. We first use DPG-GAN to generate virtual images from Gaussian noise. Then, we assign annotations to the virtual samples where person samples with the same foreground contain the same identity. Finally, we transfer the virtual person samples to the styles of different cameras with Star-GAN on average.}
\label{fig:step2}
\end{figure}

\begin{figure*}[!t]
  \centering
  \includegraphics[width=\linewidth]{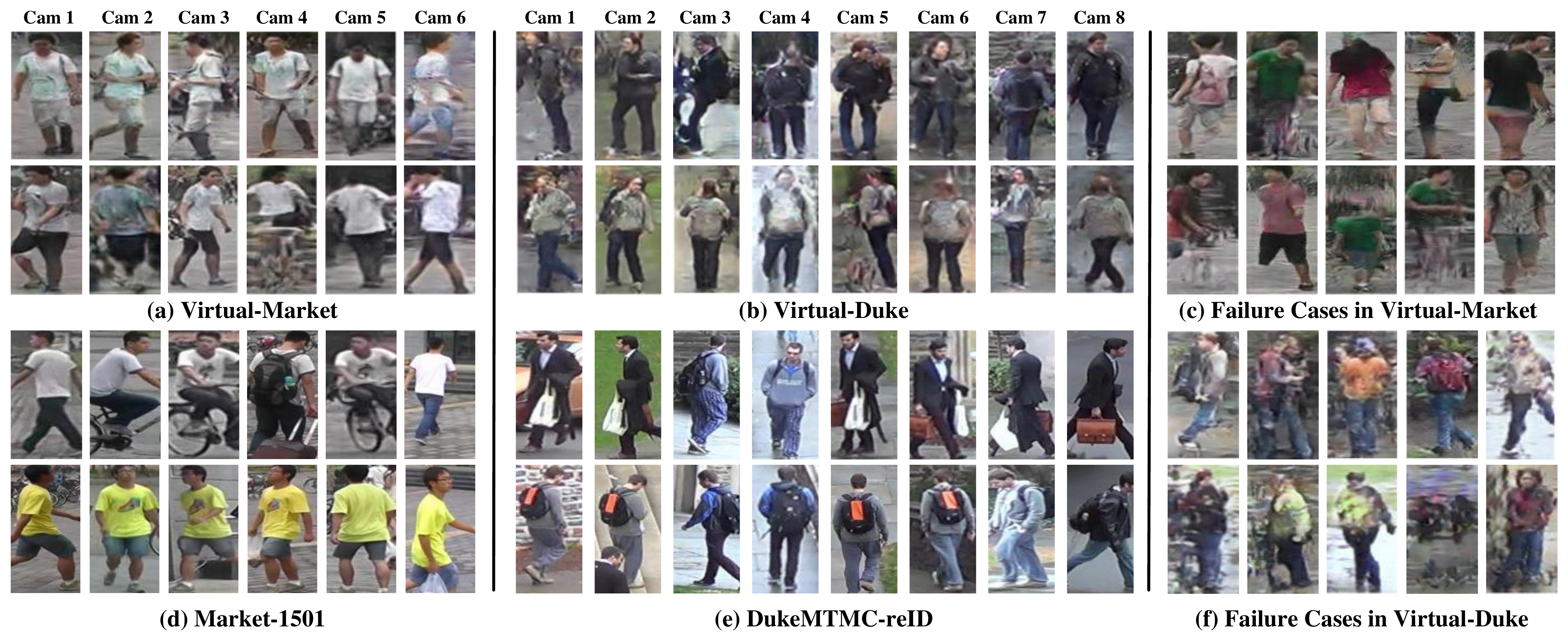}
  \caption{Examples of virtual person images on Market-1501 and DukeMTMC-reID. Despite the successful virtual images, failure instances (\textit{e.g.} incomplete body parts and blurred backgrounds) may influence the performance of deep re-ID model.}
  \label{fig:genImg}
\end{figure*}

\textbf{DPG-GAN.} DPG-GAN is an unsupervised person generation method that can obtain novel person image from Gaussian noise. A generator is proposed to disentangle pose information, foreground and background masks of unlabeled real data and encode them into embedding representations. These embeddings are decoded to reconstruct the input image with L1 loss. In addition, three generators are introduced to generate virtual embeddings from Gaussian noise and corresponding discriminators try to distinguish the embeddings of real data from the virtual embeddings. In this way, DPG-GAN leans to synthesis virtual person samples with different appearances, backgrounds and poses.

\textbf{Star-GAN.} Star-GAN consists of a style transfer model $G(x,c)$ and a discriminator $D(x)$, where $x$ and $c$ represent input image and target domain label, respectively. In this paper, we regard each camera as an independent domain. During training, $G$ is designed to generate virtual image in the style of target domain $c$. $D$ learns to distinguish between real image and style transferred image, as well as to classify the real image to its corresponding camera domain. We alternatively optimize $G$ and $D$ as the training strategy in \cite{choi2017stargan}.

\textbf{Virtual Dataset Generation.} Given unlabeled real training data, we first learn a person generation model and a camera style transfer model with DPG-GAN and Star-GAN, respectively. Then, we use DPG-GAN to randomly generate person images with different poses and transfer them in the styles of different cameras by Star-GAN. In Fig.~\ref{fig:step2}, we show the pipeline of virtual person generation, which can be summarized into four steps:

1. Define the number of identities (classes) $N_p$ and number of samples $N_e$ for each person. In this way, the number of images in the virtual dataset will be $N_p \times N_e$.

2. Sample real-like foreground $v_{fg}$, background $v_{bg}$ and pose $v_{pose}$ embeddings from Gaussian noise and feed them into pre-trained DPG-GAN for composing virtual person image. For each identity of person, we fix $v_{fg}$ and randomly sample $v_{bg}$ and $v_{pose}$ $N_e$ times to generate person images with different poses and backgrounds.

3. Repeat step 2 $N_p$ times to generate the whole virtual person dataset. Person images with the same foreground are assigned to the same identity.

4. Transfer virtual person images into styles of different cameras using pre-trained Star-GAN. For virtual person samples of each identity, we transfer them to $N_c$ camera styles on average.

To this end, we generate virtual person data with different poses and camera styles. Examples of virtual person images are shown in Fig.~\ref{fig:genImg}.

\subsection{Training Coarse Deep Re-ID Model}
\label{sec:fake-reid}
Given the labeled virtual person data contains $N_p$ identities, we are able to train a deep re-ID model as traditional methods in supervised way. In this work, we regard the training of deep re-ID model as a classification problem and train a coarse re-ID model based on IDE \cite{zheng2016person}. We adopt ResNet-50 \cite{he2016deep} as the backbone network  and add two fully convolutional (FC) layers after the Pooling-5 layer. The first FC layer has 1024-dim named as ``FC-1024''. The second FC layer named as ``FC-$\#$ID'' which has $N_p$-dim. $N_p$ is number of identities in virtual person dataset. The cross-entropy loss function is used to train the coarse re-ID model.

\subsection{Collaborative Filtering based Positive Pair Mining}
\label{sec:final-reid}
Although person generation algorithm can produce high quality samples, it still generates a certain portion of poor instances \textit{(e.g. broken limbs or blur background)} as shown in the Fig.~\ref{fig:genImg}-(c) and Fig.~\ref{fig:genImg}-(f). These poor instances will degenerate the performance of the re-ID model and the coarse deep re-Id model trained on virtual data is insufficient to discriminate the real data in testing set. To address this problem, we attempt to mine positive pairs from unlabeled data for model refinement.

\textbf{Definition.} We denote the unlabeled real data as $\mathcal{U}$. Given a query image $p \in \mathcal{U}$, our goal is to find the positive sample sharing the same identity with $p$ from $\mathcal{U}$ (except $p$). Based on the pre-trained coarse re-ID model, we extract the output of pooling-5 as the feature for each real image, and compute the pair-wise similarity matrix $\textbf{S}$ between all real images as:
\begin{equation}
 \textbf{S}_{p,q}=\exp (-||v_{p}-v_{q}||_2),
 \label{eq:Simi}
\end{equation}
where $ v_{p}$ and $ v_{q}$ are normalized pooling-5 features of image $p$ and $q$.

\textbf{$k$-reciprocal nearest neighbors.} Given the computed pair-wise similarity matrix, we could obtain the $k$-nearest neighbors (i.e. the top-$k$ samples in the similarity ranking list) for each real image. We define the $k$-nearest neighbors of $p$ as $N(p,k)$. In this paper, we adopt $k$-reciprocal nearest neighbors ($k$-RNNs) \cite{qin2011hello} instead of $k$-nearest neighbors as candidates that may contain positive samples of $p$. The $k$-RNNs for image $p$ is defined as:
\begin{equation}
    R_{k}(p) = \{q_i| (q_i \in N(p,k)) \wedge (p \in N(q_i,k)) \},
    \label{eq:krNN}
\end{equation}
where $q_i$ is among the top-$k$ similar samples of $p$, and $p$ is also among the top-$k$ of $q_i$. Intuitively, images in $R_{k}(p)$ are of high similarity with $p$ and can be utilized to form positive pairs. We named this approach as $k$-reciprocal nearest neighbor based positive pair mining. However, it will prone to form false positive pairs due to illumination, pose variation and other uncontrollable factors. To filter false samples from the candidates of $R_{k}(p)$, we then propose a collaborative filtering based positive pair mining approach to find more reliable samples that share the same identity with $p$.
\textbf{Collaborative filtering based positive pair mining.} Collaborative filtering (CF) is a technique utilized by recommender systems for preferences prediction \cite{BreeseCF}. 
The underlying assumption of the user-based CF is that if two persons have a large overlap in opinions with items, they are very likely to have a similar taste. Inspired by the user-based CF, we argue that if an image $p$ shares the same $k$-RNNs as an image $q$, they are more likely to be a positive pair. Based on the shared neighbors between $p$ and $q$, we are able to leverage their potential relations and recalculate their similarity. As shown in Fig.~\ref{fig:CF}, our approach includes four steps:

\begin{figure*}[!t]
  \centering
  \includegraphics[width=\linewidth]{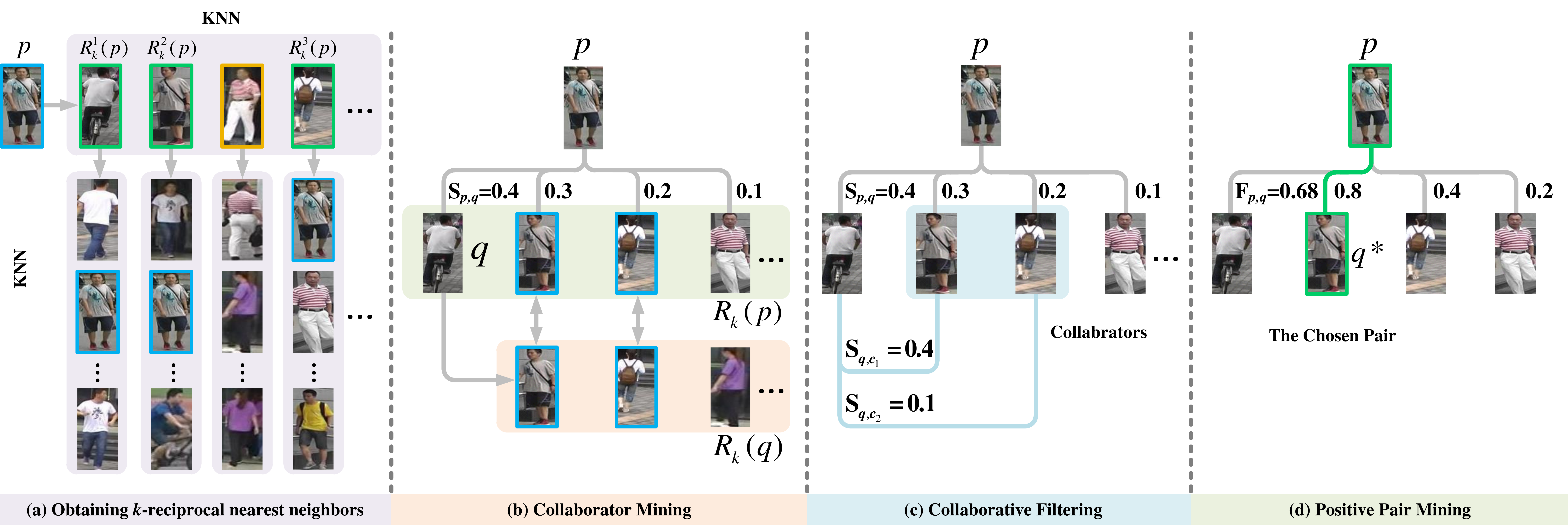}
  \caption{Collaborative filtering based positive pair mining. Given an query image $p$ (\textcolor{blue}{blue}) of real data, we first compute the $k$-reciprocal nearest neighbors $R_k(p)$ of $p$ (\textcolor{green}{green}). Then, the collaborator set (\textcolor{blue}{blue}) of $p$ and each candidate $q$ in $R_k(p)$ is mined in step (b). The collaborative filtering similarity of $p$ and each candidate $q$ in $R_k(p)$ is calculated by Eq.~\ref{eq:CF} in step (c). Finally, image pair with the highest re-calculated similarity is selected as the positive pair (\textcolor{green}{green}) in step (d).}
  \label{fig:CF}
\end{figure*}

\begin{enumerate}
\item \textit{Obtaining k-reciprocal nearest neighbors.} Given the computed pair-wise similarity matrix, we first calculate the $k$-RNNs for each real image according to Eq.\ref{eq:krNN}. For a query image $p$, we represent the $k$-RNNs of $p$ as $R_{k}(p)$ and aim to find reliable positive sample from $R_{k}(p)$.

\item \textit{Collaborator mining}. We denote collaborators as the shared $k$-RNNs of two images. Thus, given a query image $p$ and a candidate image $q$ in $R_{k}(p)$, the collaborator set $C$ of $p$ and $q$ is defined as: 

\begin{equation}
    C(p, q) = \{c_i| (c_i \in R_k(p)) \wedge (c_i \in R_k(q)) \}.
    \label{eq:CMF}
\end{equation}

\item \textit{Collaborative filtering similarity}. Based on the collaborator set of $p$ and $q$, we calculate the filtered similarity as:

\begin{equation}
    \textbf{F}_{p,q} = \textbf{S}_{p,q} + \sum_{i=1}^{|C|}{\textbf{w}_{q,c_i}\textbf{S}_{p,c_i}},
    \label{eq:CF}
\end{equation}
where $| \cdot |$ denotes number of candidates in a set, and $\textbf{w}_{q,c_i}$ is the normalized weight to measure the significance of collaborator $c_i$, defined as:
\begin{equation}
    \textbf{w}_{q,c_i} = \frac{\textbf{S}_{q,c_i}}{\sum_{i=1}^{|C|}{\textbf{S}_{q,c_i}}}.
    \label{eq:normalize}
\end{equation}
The filtered similarity not only considers the original pair-wise distance of $p$ and $q$, but also takes the similarities between $p$, $q$ and the collaborator set into consideration.

\item \textit{Positive pair mining}. With the calculated  collaborative filtering similarities between query image $p$ and images in $R_k(p)$, image $q^*$ with the highest similarity $F$ is selected to construct a positive pair $(p, q^*)$ for re-ID model fine-tuning:

\begin{equation}
    q^* = \mathop{\arg\max}_{q \in R_k(p)} \textbf{F}_{p,q}.
    \label{eq:PPM}
\end{equation}

In practice, we find that positive pairs obtained by our algorithm are always in the same camera. This may make the re-ID model sensitive to camera variations, while the main goal of re-ID is to retrieval a person across different cameras. To alleviate this problem, we remove all images that share the same camera with the given image $p$ during the calculation of $k$-RNNs.
\end{enumerate}

\begin{table*}[t]
    \centering
    \caption{Unsupervised person re-ID performance comparison with state-of-the-art methods on Market-1501 and DukeMTMC-reID.}
    \label{tab:cmpDA}
    \resizebox{\textwidth}{!}{
    \begin{tabular}{c|ccccc|ccccc} \hline
        \multirow{1}{*}{Methods} & \multicolumn{5}{c}{Duke $\rightarrow$ Market} \vline & \multicolumn{5}{c}{Market $\rightarrow$ Duke} \\
        \cline{2-11}
         (Domain Adaptation) & mAP & rank-1 & rank-5 & rank-10 & rank-20 & mAP & rank-1 & rank-5 & rank-10 & rank-20 \\
        \hline
        UMDL \cite{peng2016unsupervised} & 12.4 & 34.5 & 52.6 & 59.6 & - & 7.3 & 18.5 & 31.4 & 37.6 & -\\
        PT-GAN \cite{wei2017person} & - & 38.6 & - & 66.1 & - & - & 27.4 & - & 50.7 & -\\
        SP-GAN \cite{deng2018image} & 22.8 & 51.5 & 70.1 & 76.8 & - & 22.3 & 41.1 & 56.6 & 63.0 & -\\
        \hline
        {Methods} & \multicolumn{5}{c}{Market-1501} \vline & \multicolumn{5}{c}{DukeMTMC-reID} \\
        \cline{2-11}
        (Unsupervised) & mAP & rank-1 & rank-5 & rank-10 & rank-20 & mAP & rank-1 & rank-5 & rank-10 & rank-20 \\
        \hline
        LOMO \cite{liao2015person} & 8.0 & 27.2 & 41.6 & 49.1 & - & 4.8 & 12.3 & 21.3 & 26.6 & -\\
        Bow \cite{zheng2015scalable} & 14.8 & 35.8 & 52.4 & 60.3 & - & 8.3 & 17.1 & 28.8 & 34.9 & -\\
        DPG-GAN \cite{ma2018disentangled} & 13.8 & 33.8 & - & - & - & 9.0 & 19.5 & 33.3 & 39.9 & 47.9\\
        UPL \cite{fan2017unsupervised} & 20.1 & 44.7 & 59.1 & 65.6 & 71.7 & 16.4 & 30.4 & 44.5 & 50.7 & 56.0\\   
        CAMEL \cite{camel} & 26.3 & 54.5 & - & - & - & - & - & - & - & -\\
        \hline
        Our Method  & \textbf{33.9} & \textbf{63.9} & \textbf{81.1} & \textbf{86.4} & \textbf{90.8} & \textbf{17.9} & \textbf{36.3} & \textbf{54.0} & \textbf{61.6} & \textbf{67.8}\\
        \hline
      \end{tabular}}
\end{table*} 

\subsection{Model Fine-tuning}
\label{sec:model-finetune}

After mining the positive pairs of real data, we combine them together with the generated virtual data to refine the previous coarse deep Re-ID model. Triplet loss project similar pairs into a feature space with a smaller distance than dissimilar pairs, which can be adopted for selected positive training pairs. Another reason to use triplet loss on positive pairs is that we do not have the real label for selected real images, cross-entropy loss can not be obtained.

During training, we randomly sample $N_r$ anchor images from real data and their corresponding mined positive samples to form the training batch. For each image $p_i$, we assign the same pseudo class to $p_i$ and its mined positive sample $q_i^*$. We select the hardest (closest) sample $z_i$ from the other $N-2$ images as the negative sample of $p_i$. The final triple loss function is as following, 
\begin{equation}
  \begin{aligned}
      L_{tri}=\sum_{i=1}^{N_r}\Big[||f(p_i)-f(q_i^*)||_2- ||f(p_i)-f(z_i)||_2+m\Big]_{+},
  \end{aligned}
  \label{eq:triLoss}
\end{equation}
where $m$ is a margin that is enforced between positive and negative pairs, and $f(\cdot)$ is the pooling-5 feature of the deep re-ID model. $N_r$ is the number of anchors in the training batch.

As we already have the pseudo labels of the generate virtual data, we directly use the IDE cross-entropy loss function $L_{cls}$. By merging these two losses into a multi-task training framework, we then have the final loss as:
  \begin{equation}
    L_{loss}=L_{cls}+\lambda L_{tri},
    \label{eq:lossStage2}
  \end{equation}
where $\lambda$ is a hyper-parameter controlling the influence of $L_{cls}$ and $L_{tri}$.

When finished training the re-ID model for each epoch, the parameters of the deep re-ID model will be updated and the adjacent matrix $\textbf{S}$ of the real data will also be updated. As a result, we need to proceed a positive pair mining step for each epoch. The final model can be trained by using loss function~\ref{eq:lossStage2}. By doing so, the real data can help increase the final re-ID accuracy by eliminating negative effects of distorted virtual images while virtual data stabilizes the training process and the keep basic performance of re-ID model. 

\section{Experiments}
To evaluate the performance of our proposed method, we conduct experiments on two large-scale benchmark datasets: Market-1501~\cite{zheng2015scalable} and  DukeMTMC-reID~\cite{zheng2017unlabeled,ristani2016MTMC}. 
The mAP and rank-1 accuracy are adopted as evaluation metrics.

\textbf{Market-1501} dataset contains 32,668 bounding boxes of 1,501 identities obtained from six cameras. 
751 persons are used for training while the rest for testing (750 identities, 19,732 images). The probe set contains 3,338 images for querying true person images from gallery set.

\textbf{DukeMTMC-reID} dataset is a subset of DukeMTMC \cite{ristani2016MTMC} which consists of 36,411 labeled bounding boxes of 1,404 identities pictured by 8 different cameras. Similar to the protocol of Market-1501, this dataset split 16,522 images of 702 identities for training, 2,228 probe images and 17,661 gallery images from the rest for testing.

\subsection{Experiment Settings}
\label{sec:exp}
\textbf{DPG-GAN:} We train the DPG-GAN by 120,000 epochs with a batch size of 16. The learning rates of all networks to 0.00008 and divided by 10 in every 10,000 epochs. All input images are resized to 128$\times$64. We use the same network architectures following \cite{ma2018disentangled}. 

In virtual person generation stage, we use $N_p$ to represent the number of individuals/identities included in virtual dataset while $N_e$ denotes the number of images generated for each person. Unless otherwise specified, we generate virtual datasets with $N_p=600$ and $N_e=36$ for Market-1501, and with $N_p=600$ and $N_e=48$ for DukeMTMC-reID.

\textbf{Star-GAN:} The Adam solver is employed to train $G$ and $D$ of Star-GAN for a total 200 epochs with a batch-size of 40. Input images are resized to 128$\times$128. The learning rates for $D$ and $G$ are initialized to 0.0001 and linearly reduced to 0 for the last 100 epochs. We employ the network structures following \cite{choi2017stargan}.

During camera style translation, one-hot label of target camera is tiled and concatenated with input images to form a 128$\times$128$\times$($N_c$+3) tensor, the tensor is then sent to U-Net-like generator for style translation. $N_c$ is the total number of cameras for corresponding real dataset. We convert images from virtual data to different camera styles on average. In other words, each image is transferred to one style of cameras.

\textbf{Re-ID model training:} We resize input image to 256$\times$128, and employ random horizontal flipping and random cropping for data argumentation. The SGD solver is used for optimization with a learning rate initialized as 0.1 and divided by 10 after 100 epochs. We train the re-ID model with 150 epochs in total. For positive pair mining, we first train the re-ID model by only using the virtual data for 100 epochs. After that, we add the mined positive pairs from real data for fine-tuning with another 50 epochs. Other parameters are set as follows: the triplet loss with anchor batch size $N_r=50$ and a margin $m=0.3$, $k$-reciprocal nearest neighbors with $k=50$, and $\lambda=1$ in Eq.~\ref{eq:lossStage2}.

\subsection{Comparison with State-of-The-Art}
In order to compare with other competing unsupervised re-ID methods, we train two models with generated virtual datasets for the Market-1501 and DukeMTMC-reID dataset, respectively. 
All the experimental results of our method and other methods are reported in Table~\ref{tab:cmpDA}. As can be seen, our method outperforms all previous unsupervised re-ID methods. On Market-1501, we can get a rank-1 accuracy of 63.9\%, which is 9.4\% higher than the previous state-of-the-art method CAMEL~\cite{camel}. On DukeMTMC-reID, our method also can beat UPL~\cite{fan2017unsupervised} with a 5.9\% higher rank-1 accuracy.
Additionally, we also compare our method with three domain adaptation based methods. Domain adaptation methods can be considered as semi-supervised which training a model from one labeled dataset and then transfer to another different datasets. As can be seen from Table~\ref{tab:cmpDA}, our method outperform all three methods on Market-1501 dataset, with a 12.4\% higher rank-1 accuracy compared with the best SP-GAN. On the DukeMTMC-reID dataset, the accuracy of our method is higher than the UMDL and PT-GAN, but lower than the SP-GAN. The main reason is that the generated virtual images still contain lots of low-quality samples which directly affect the accuracy of our method.

\subsection{Ablation Study}

\begin{table}[t]
    \centering
    \caption{Ablation study of our approach. Based on the re-ID model trained on DPG-GAN, we add Star-GAN, positive pair mining gradually into it to evaluate the re-ID accuracy.}
    \label{tab:graduallyCMP}
\footnotesize
\begin{tabular}{l|cc|cc} 
        \hline
       \multirow{2}{*}{Method}  &  \multicolumn{2}{c|}{Market-1501}  & \multicolumn{2}{c}{DukeMTMC-reID} \\
          & mAP & rank-1  & mAP & rank-1 \\
        \hline
        DPG-GAN & 13.4 & 33.8  & 9.0 & 19.5\\
        \hline
        + Star-GAN & 25.1 & 51.7  & 13.9 & 30.3\\
        \hline
        + Star-GAN+Mining & \textbf{33.9} & \textbf{63.9} & \textbf{17.9} & \textbf{36.3}\\
        \hline
    \end{tabular}
\end{table}

\begin{figure}[!t]
    \centering\includegraphics[width=\linewidth]{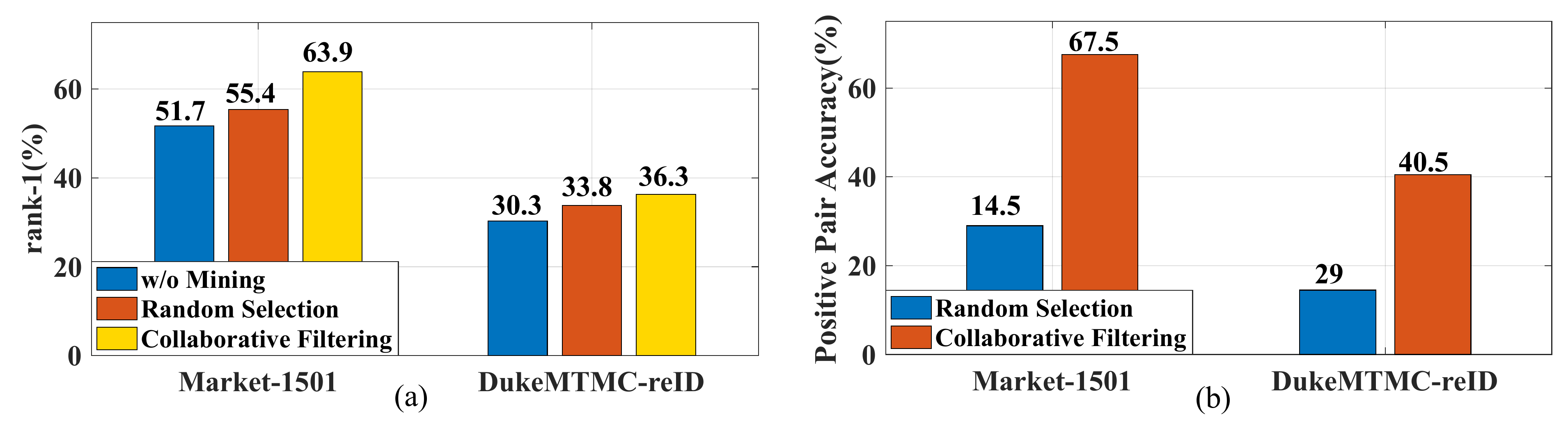}
    \caption{The effectiveness of collaborative filtering based positive pair mining. We compare with the re-ID models trained without mining and with random selection mining.}
    \label{fig:randCF}
\end{figure}

The method discussed in Section~\ref{sec:model} contains three main components: DPG-GAN, Star-GAN and positive pair mining. In order to figure out which component contributes the most for the accuracy, we evaluate the performance by gradually add the Star-GAN and positive pair mining into the re-ID model training. As can be seen in Table~\ref{tab:graduallyCMP}, after adding the Star-GAN, the rank-1 on Market-1501 dataset can boost from 33.8\% to 51.7\%, which means camera-style transfer each generated identity across different cameras playing a significant role for re-ID model initialization.  Then including the positive pair mining step, we observe a further 10.3\% rank-1 improvement on Market-1501. Adding real data into training can help reducing the gap between the generate virtual data and real data. On the DukeMTMC-reID dataset, we have similar findings. 
 
To assess the effectiveness of proposed collaborative filtering based mining procedure, we perform another comparison between \textit{without mining}, \textit{random selection} and \textit{collaborative filtering}. Random selection is simplified mining procedure which randomly chooses samples from the $k$-reciprocal nearest neighbors of anchor images as positive pairs. As shown in Fig.~\ref{fig:randCF}-(a), the mining step can improve the result by a large margin, and the proposed collaborative filtering based mining outperform random selection on rank-1 by 6.6\% and 2.9\% on two datasets.

In triplet training phase. We randomly select $N$ anchor images and their corresponding mined positive samples to form the training batch. For each anchor, we randomly select hardest sample as the negative within the other $N-2$ images and their corresponding positive samples. In this way, the probablity of selecting the real positive sample as the negative is very low when sampling a few images from a dataset containing a large number of images and identities. We evaluate the overall error rate during Market-1501 and DukeMTMC-reID training. They are 2.6\% and 3.1\%, respectively. Due to our competitive performance, we would consider the error rate to be small, so it is not necessary to avoid selecting the real positive sample as the negative. 

We also validate the accuracy of the mined pairs belonging to the same identity during the whole fine-tuning step by using the ground-truth information of two datasets in Figure~\ref{fig:CF}-(b). As shown, the accuracy of random selection is only 29.0\% and 14.5\% for Market-1501 and DukeMTMC-reID. After employing the collaborative filtering, the accuracies rise to 67.5\% and 40.5\%, which means the quality of mined positive pairs directly affect the final performance.

\subsection{Sensitive Analysis}
\label{sec:SA}

To check the sensitive of method with different hype-parameters, we do a thorough evaluation of: (1) the scale of generated virtual dataset ($N_p$ and $N_e$), (2) the batch-size of real positive pair $N_r$ and the value of $k$, (3) the influence of the $\lambda$ for the $L_{cls}$ and $L_{tri}$.

\begin{figure}[t]
  \centering
  \includegraphics[width=\linewidth]{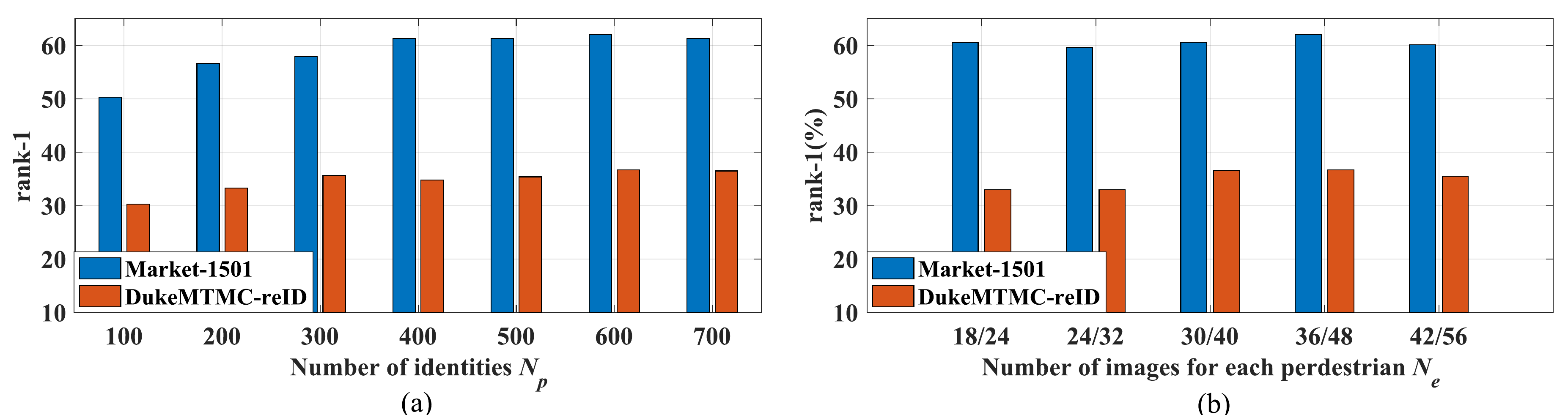}
  \caption{Sensitive analysis for $N_e$ and $N_p$. Increasing the size of virtual dataset may help improve re-ID accuracy in a certain degree.}
  \label{fig:nSA}
\end{figure}

\textbf{Large-scale virtual dataset has positive impact}. Intuitively, we conduct a series experiments to evaluate the influence of the scale of the generated virtual datasets. Figure~\ref{fig:nSA}-(a) presents the relationship between $N_p$ and rank-1, by changing $N_p$ from 100 to 700 with a fixed $N_e$ equals to 36 and 48 for Market-1501 and DukeMTMC-reID. In general, re-ID model performs better with larger $N_e$, but the accuracy will begin to saturate when $N_e$ is large enough. The same is true for $N_e$ as shown in Figure~\ref{fig:nSA}-(b). We fix $N_p$ to 600 and vary $N_e$. The rank-1 accuracy will saturate when $N_e=36$ for Market-1501 and 48 for DukeMTMC-reID, respectively. The results demonstrate that enlarging the scale of virtual dataset can improve performance of model in a certain degree.
  
\begin{figure}[!t]
  \centering\includegraphics[width=\linewidth]{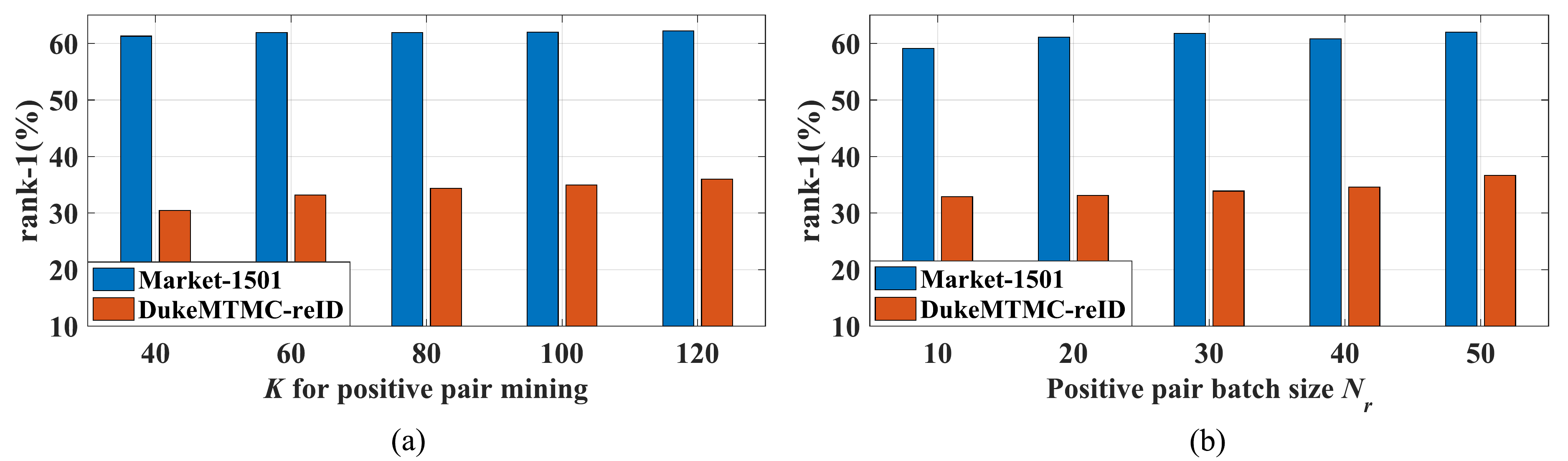}
  \caption{Sensitive analysis for $k$ and $N_r$. Our approach is robust to the changes of $k$ and $N_r$.}
  \label{fig:kSA}
\end{figure}

\begin{figure}[!t]
  \centering\includegraphics[width=\linewidth]{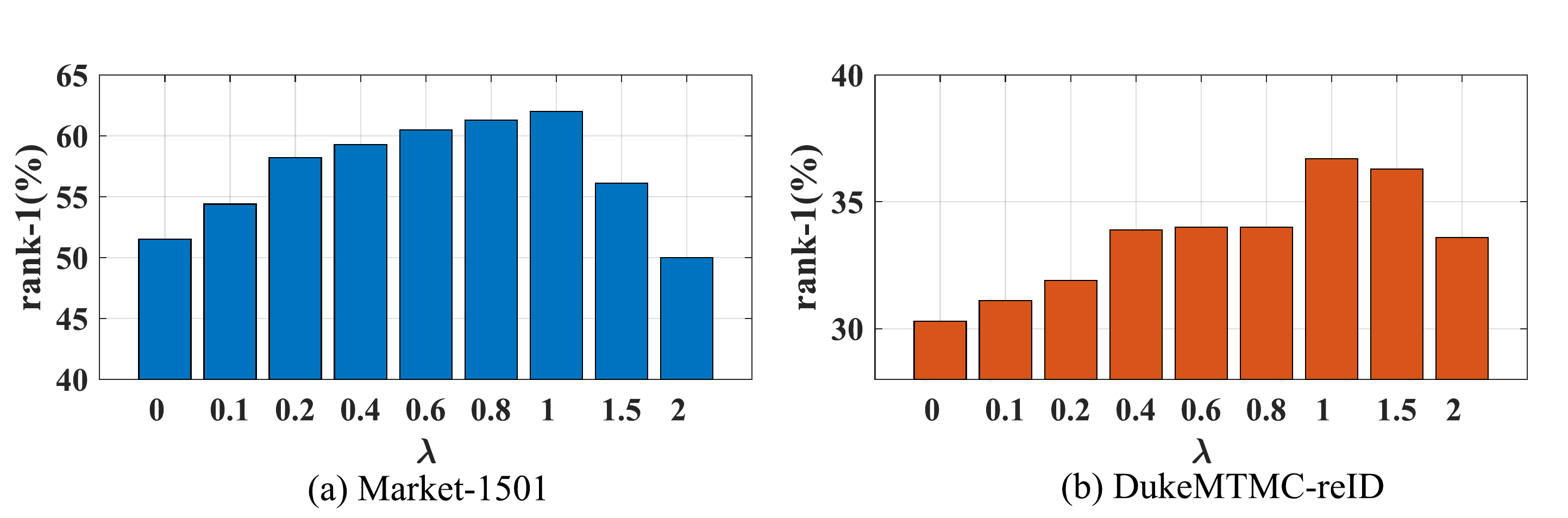}
  \caption{Sensitive analysis for $\lambda$ shows that $L_{tri}$ and $L_{cls}$ contribute equally to the  accuracy of PGPPM. Best result is achieved when $\lambda$ is around 1.}
  \label{fig:lam}
\end{figure}

\textbf{Various $N_r$ and $k$ have less effect.} We perform another experiment to check how many positive pairs are needed for fine-tuning the final re-ID model. As shown in Figure~\ref{fig:kSA}-(a) and (b), the rank-1 accuracy are fluctuated in a very small range around 63.9\% and 36.3\% for Market-1501 and DukeMTMC-reID by using various $N_r$ and $k$. But in practice, we still suggest using a large $k$ to ensure that $k$-reciprocal nearest neighbors can always be found.

\textbf{Both $L_{cls}$ and $L_{tri}$ are important for model optimization.} We evaluate PGPPM model with different $\lambda$ values to find out which part contributes most for the accuracy of model. Figure~\ref{fig:lam} shows that rank-1 accuracy of PGPPM improves with the increase of $\lambda$ when $\lambda$ is in the range of [0,1]. However, when $\lambda$ exceeds 1, the rank-1 score begins to decrease. The best result is achieved when $\lambda$ is around 1. The results prove our claims that both $L_{cls}$ and $L_{tri}$ are important for our model. $L_{cls}$ helps model to learn robust features while $L_{tri}$ eliminates the negative effects brought by virtual images.

\section{Conclusion}
In this paper, we consider a challenging problem in person re-identification (re-ID), where labels are not provided in training data. To optimize deep re-ID model in supervised way, this work generates virtual dataset with a person generation model and a camera style model. Moreover, a collaborative filtering based positive pair mining approach is proposed to find reliable positive sample for refining re-ID model. Experiments on two benchmark datasets show that our method outperforms current unsupervised re-ID algorithms. In the future work, we will focus on learning a person generation model that jointly considers the pose and camera variations and produces higher quality virtual images.

\bibliographystyle{IEEEtran}
\bibliography{IEEEabrv,IEEEexample}

%








\end{document}